\documentclass[letterpaper]{article} 
\usepackage{aaai24}  
\usepackage{times}  
\usepackage{helvet}  
\usepackage{courier}  
\usepackage[hyphens]{url}  
\usepackage{graphicx} 
\usepackage{natbib}  
\usepackage{caption} 
\usepackage{algorithm}
\usepackage{algorithmic}

\usepackage{newfloat}
\usepackage{listings}
\DeclareCaptionStyle{ruled}{labelfont=normalfont,labelsep=colon,strut=off} 
\floatstyle{ruled}
\newfloat{listing}{tb}{lst}{}
\floatname{listing}{Listing}
\usepackage{pifont}
\usepackage{booktabs}
\usepackage{amssymb}
\usepackage{amsmath}
\usepackage{multirow}
\usepackage{multicol}
\usepackage{tabularx}
\usepackage{adjustbox}
\usepackage{xcolor}
\usepackage{anyfontsize}
\newcommand{\xmark}{\ding{55}}
\newcommand{\cmark}{\ding{51}}

\title{Gradient-Guided Modality Decoupling for Missing-Modality Robustness}
\author{
    Hao Wang\textsuperscript{\rm 1},
    Shengda Luo\textsuperscript{\rm 1},
    Guosheng Hu\textsuperscript{\rm 2},
    Jianguo Zhang\textsuperscript{\rm 1,3}\thanks{Corresponding author}
}
\affiliations{
    \textsuperscript{\rm 1}Research Institute of Trustworthy Autonomous Systems and Department of Computer Science and Engineering, Southern University of Science and Technology, Shenzhen 518055, China\\
    \textsuperscript{\rm 2}Oosto, BT1 2BE, Belfast, U.K.\\

    \textsuperscript{\rm 3}Peng Cheng Lab, Shenzhen, China\\
    12232399@mail.sustech.edu.cn, shengdaluo1991@gmail.com,\\ huguosheng100@gmail.com, zhangjg@sustech.edu.cn

}

\usepackage{bibentry}

\begin{document}

\maketitle

\begin{abstract}
Multimodal learning with \textit{incomplete} input data (missing modality) is practical and challenging. In this work, we conduct an in-depth analysis of this challenge and find that modality dominance has a significant negative impact on the model training, greatly degrading the missing modality performance. Motivated by Grad-CAM, we introduce a novel indicator, gradients, to monitor and reduce modality dominance which widely exists in the missing-modality scenario. In aid of this indicator, we present a novel Gradient-guided Modality Decoupling (GMD) method to decouple the dependency on dominating modalities. Specifically, GMD removes the conflicted gradient components from different modalities to achieve this decoupling, significantly improving the performance. In addition, to flexibly handle modal-incomplete data, we design a parameter-efficient Dynamic Sharing (DS) framework which can adaptively switch on/off the network parameters based on whether one modality is available. We conduct extensive experiments on three popular multimodal benchmarks, including BraTS 2018 for medical segmentation, CMU-MOSI, and CMU-MOSEI for sentiment analysis. The results show that our method can significantly outperform the competitors, showing the effectiveness of the proposed solutions. Our code is released here: https://github.com/HaoWang420/Gradient-guided-Modality-Decoupling.
\end{abstract}

\section{Introduction}
\label{sec:intro}
%

Multimodal learning \cite{ma2022multimodal, tsai2019multimodal}, in contrary to traditional tasks performed on a single modality, aims to utilize multiple sources of information from distinct modalities. 
Various multimodal frameworks have been proposed to fully exploit complementary information across multiple modalities. 
{In the real world, it is not always easy or feasible to obtain the data of all the modalities, leading to practical research to address this missing-modality problem. }
In particular, in the medical domain, missing modality is a vital problem. 
As a matter of routine, healthcare clinics and facilities may lack the necessary infrastructure to obtain valuable data of all the modalities, leading to a shortage of critical diagnostic information. 
\begin{table}[t]
\centering
\fontsize{9pt}{9pt}\selectfont
\centering
\begin{tabular}{l|ccc|c|c}
\toprule
                  & \multicolumn{3}{c|}{Unimodal}                 & Multimodal      & \multicolumn{1}{l}{} \\ 
                  & \multicolumn{3}{c|}{(missing 3)}                 & (missing 1)      & \multicolumn{1}{l}{} \\ \midrule
Method            & Flair         & T2            & T1            & Flair,T2,T1 & $\Delta$ (M.-U.)   \\ \midrule
U. base. & 58.3          & 56.3          & 71.0          & -               & -                    \\
ACN               & 67.7          & 67.9          & 71.2          & 67.9            & -3.3                 \\
SMU-N.           & 71.8          & 67.2          & 69.5          & 67.9            & -3.9                 \\
RFNet             & 69.2          & 71.0          & \textit{66.0}          & 75.2            & +4.1                 \\
\textbf{Ours }             & \textbf{74.1} & \textbf{75.4} & \textbf{73.2} & \textbf{79.0}   & \textbf{+4.6}  \\     
\bottomrule
\end{tabular}
\caption{
Two scenarios, unimodal (missing 3 modalities) and multimodal (misssing 1 modality) evaluated on BraTS 2018 TC task with DSC ($\uparrow$) score. Existing methods either exhibit lower multimodal than unimodal performance (ACN, SMU-Net) or have worse unimodal performance  than baseline (T1 of RFNet), indicating unresolved modality dominance. In contrast, our method addresses this problem and achieves the best performance. 
}
\label{tab:dominance}
\end{table}
Therefore, improving the robustness of multimodal algorithms against modal-incomplete data is crucial for applying deep learning in various multimodal tasks. Additionally, humans, unlike multimodal systems, possess a remarkable capability of recognizing concepts, objects, and sentiments even when presented with only partial modalities or missing senses. For example, a person can still effectively understand the emotional state of another by listening to their tone of voice, even without visual access to their facial expressions. This inherent robustness against missing modalities in human perception serves as part of the motivation for us to improve the robustness of multimodal models.

Existing methods for handling missing-modality problems can be roughly classified into three categories:
(1) distillation and co-training between multiple networks; (2) reconstruction-based modal-completion; (3) missing-modality robust architecture.
For distillation or co-training approaches \cite{xing2022kd, wang2021acn, poklukar2022geometric}, 
Although they marginally improve the missing-modality robustness, their methods are highly dependent on the choice of similarity measurements and require modal-incomplete input to be filled with masking values \cite{wang2021acn,tsai2019multimodal},
which may cause unexpected behavior in the models, leading to degraded performance \cite{shen2019brain}.
Reconstruction-based methods \cite{khattar2019mvae, shi2019variational, suttergeneralized, Zhao2021mmin, Lian2022GCNetGC} 
unavoidably bring high computational costs and excessively rely on the quality of reconstructed samples. 
In contrast to the reconstruction-based method, Ma et al. \cite{ma2022multimodal} proposed an architecture based on Transformer models to improve missing-modality robustness via special attention masks and dynamic fusion strategies. Their method is limited to transformer-based architectures. {
Despite the success of these methods, they do not intrinsically investigate 
the relationship among various modalities.}

Under the missing-modality setting, we find that the \textit{modality dominance} problem \cite{wu2022characterizing, wang2020what, huang2022modality, peng2022balanced}, which means multimodal models tend to rely on only the dominating modality while under-fitting the other modalities, can  particularly degrade the performance. 
First, it leads to under-optimization of unimodal ones, e.g. T1 of RFNet \cite{ding2021rfnet} works worse than baseline in Table \ref{tab:dominance}. 
Second, modality dominance suppresses the capability of modalities in multimodal settings.
For example,  ACN and SMU-Net have weaker multimodal  performance than the unimodal one in Table \ref{tab:dominance}. 
This indicates the models are overly dependent on specific modalities and unable to properly integrate information from other modalities. Both cases significantly affect robustness to missing modalities. 

To reduce modality dominance and improve the missing-modality robustness, we propose a novel gradient-guided approach, which investigates the relations of different modalities by analyzing their gradients. 
Motivated by Grad-CAM \cite{selvaraju2017grad}, where gradients can quantify the contributions of different components in a model,  in the multimodal learning context, we propose that gradients of different modalities can also indicate their relative significance and dependency relationships.
With the aid of gradients, we achieve new insights into modality dominance, which can be interpreted as gradient conflicts. Specifically, we find modality dominance occurs when gradients from different modalities possess varied norms and opposing directions that cancel each other out. {This finding inspires us to address the missing modality problem by adjusting and balancing the gradients to ensure that all modalities are well represented during training, thereby preventing any single modality from dominating the others.} 
Unlike OGM-GE \cite{peng2022balanced}, which modulates the gradients by their contribution to the objective while ignoring inter-modality relation, our method focuses on addressing the conflicts between modalities and improving the robustness of multimodal models.

In this work, we propose a Gradient-guided Modality Decoupling (GMD) method to balance modalities by analyzing the gradients. Specifically, GMD  decouples the entangled gradients of different modalities by cancelling the conflicting gradients and keeping the modality-specific gradients, aiming to reduce modality dominance. 
Furthermore, we design a Dynamic Sharing (DS) technique that flexibly reduces the impact of absent modalities. 
Presented with modal-incomplete input, DS does not require input to be completed by filling 0s like many methods \cite{wang2021acn,azad2022smu} did.
Instead, it can adaptively switch off the 
network parameters corresponding to the missing modalities, avoiding introducing misleading 0 fillers to the network.

Our contribution can be summarized as follows:
\begin{itemize}
    \item {We conduct an in-depth analysis of the missing modality problem and find that \textit{modality dominance} substantially limits performance. This finding inspires us to address this missing modality problem by balancing the modalities. We adopt gradients as the effective indicators to measure the importance of modalities, leading to the solution of resolving modality dominance by balancing the gradients.} 
    \item We propose a Gradient-guided Modality Decoupling (GMD) method, which resolves modality dominance by removing \textit{conflicted gradient components}. It significantly improves the robustness against the missing modality problem. 
    \item 
    Unlike many existing methods, which add masking values (e.g., 0s) in the image or feature space to impute missing modalities, 
    we present a simple yet parameter-efficient Dynamic Sharing (DS) architecture, which can adaptively switch off the network parameters related to the missing modalities. 
    \item {We conduct extensive experiments on the medical segmentation dataset BraTS 2018, two sentiment analysis datasets, CMU-MOSI and CMU-MOSEI. Our method achieves state-of-the-art performance on these datasets, demonstrating the great effectiveness of our approach.}
\end{itemize}
\section{Related Work}
\paragraph{Multimodal learning}  In multimodal learning, different modalities, e.g., texts, images \cite{zhang2023decoding}, videos \cite{zadeh2016mosi, zadeh2018mosei}, Magnetic Resonance Image (MRI) modalities \cite{MRI2021Dinsdale} or other sources of information \cite{Lu2022ActionconditionedOM, Zhang2022CharacterizingPH}, are often considered to be complementary to each other. 
Fusion has been widely studied as one of the most critical topics in multimodal learning.
Nagrani et al. explored fusion strategies and proposed the attention bottleneck for fusing various modalities \cite{nagrani2021attention}.

 
\paragraph{Transfer knowledge for missing modality robustness}
Recent studies \cite{latent, shen2019brain, wang2021acn, azad2022smu} have introduced Knowledge Distillation \cite{gou2021knowledge, xing2022kd} or co-training \cite{nigam2000analyzing} for improving missing-modality robustness. 
Despite the computational and memory cost of additional networks in their method, the choice of similarity measurements  significantly impacts the final performance \cite{azad2022smu}. 
Moreover, their methods still require missing-modality data to be completed with masking values, which may cause unexpected behavior in the models \cite{shen2019brain}.

\paragraph{Reconstruction-based modality completion}  To effectively complete the absent modalities, some \cite{khattar2019mvae, Zhao2021mmin, Zhang2019CPMNetsCP, Lian2022GCNetGC} propose to introduce reconstruction-based methods, e.g., Variational Auto-Encoders (VAEs) \cite{shi2019variational} or Generative Adversarial Networks (GANs) \cite{zhou2021conditional}. MVAE \cite{khattar2019mvae} adopts a Product-of-Expert (PoE) inference network to learn a joint-modality representation. MMVAE \cite{shi2019variational} partially addressed the over-confident problem of MVAE with a computationally expensive gradient estimator. MoPoE \cite{suttergeneralized} provides a generalized multimodal ELBO by combining MVAE and MMVAE.
Intuitively, reconstruction can improve robustness by completing the missing modalities.
However, the final performance heavily depends on the reconstruction quality.

\paragraph{Modality dominance}
Recent works on multimodal learning \cite{wu2022characterizing, wang2020what, peng2022balanced, huang2022modality} empirically studied the vast existence of modality dominance, which indicates the model solely relies on one critical modality, yet other modalities remain under-optimized. 
OGM-GE \cite{peng2022balanced} approaches the modality dominance problem from the optimization perspective \cite{yu2020gradient}, which modulates the gradients by their contribution to the objective. However, OGM-GE ignores the inter-modality relation and is limited to bi-modal problems. 
In our work, we reformulate the modality dominance incurred by the dependency of models on dominating modalities into gradient conflicts. From the optimization perspective, we propose a method to resolve the modality dominance with the guidance of gradient. 

\begin{figure*}[ht]
    \centering
    \includegraphics[width=0.9\textwidth, clip, trim=0cm 0cm 0cm 0cm]{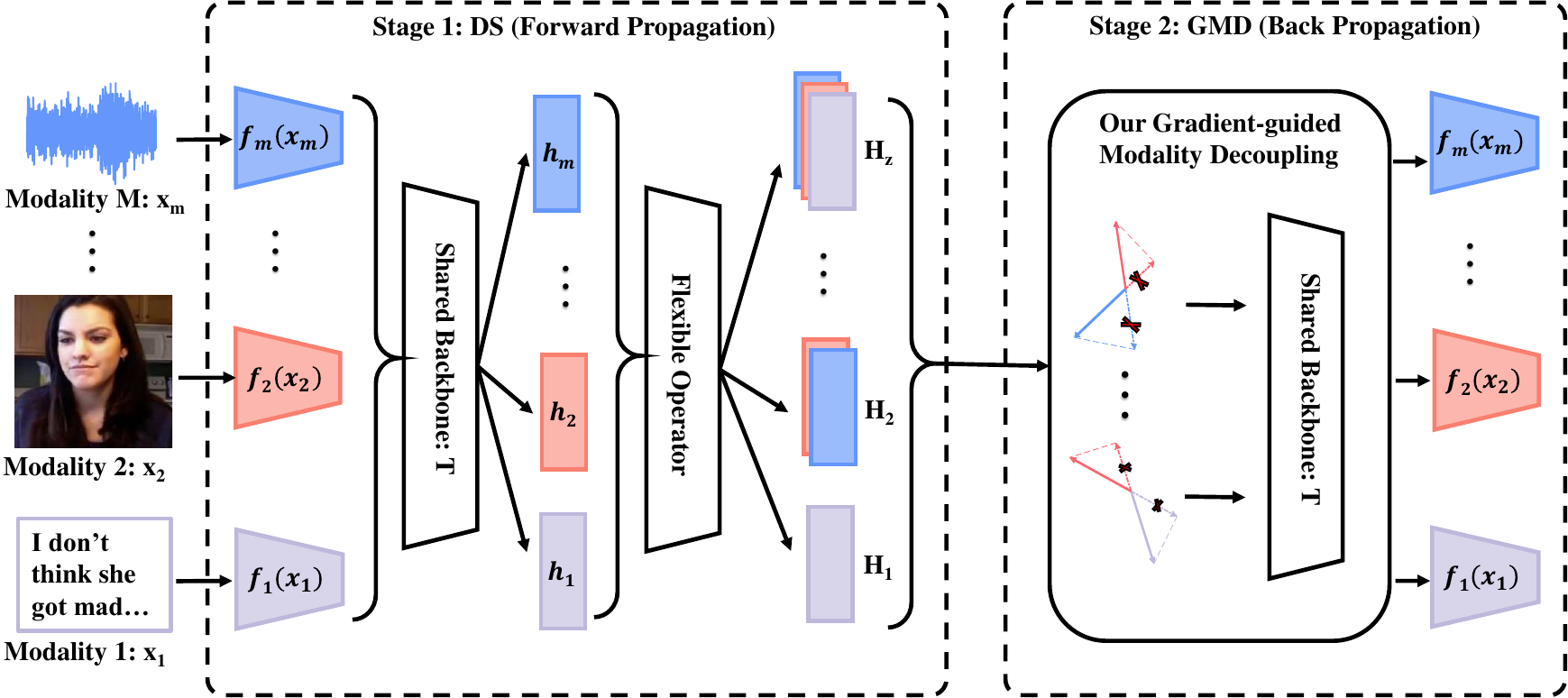}
    \caption{{
        Overall architecture including Dynamic Sharing (DS) and Gradient-guided Modality Decoupling (GMD). Multimodal data is first mapped to a common feature space through modality encoders $\{f_1, f_2,...,f_m\}$. 
    Then the outputs of these encoders are fed to a shared architecture: $T$. Then the outputs $\{h_1, h_2,...,h_m\}$ go through a flexible fusion operator (e.g., average pooling) to obtain modal-incomplete representations $\{H_1, H_2,..., H_z\}$. GMD  decouples the conflicted gradients, and the calibrated gradients are back-propagated to reduce modality dominance. }}
    \label{fig:ds_framework}
\end{figure*}

\section{Method}


\subsection{Overall Architecture}
\label{Overview}
Figure \ref{fig:ds_framework} presents an overview of our architecture. Let $\{X, Y\}$ represents a sample from a multimodal dataset where $X = \{x_1, x_2, \dots, x_m\}$  with $m$ modalities and $Y$ is the corresponding label. 
In our framework, the data from various modalities are first mapped to a common feature space via modality encoders $\{f_1, f_2, ..., f_m\}$.
A shared backbone $T(\cdot; \theta)$ parameterized by $\theta$ is used to process features separately to obtain the unimodal representation $h_i$ for each modality.
\begin{equation}
    h_i = T(f_i(x_i); \theta)
\end{equation}
At the end of the shared backbone, the mapped representations are fused via flexible operator $\mathcal{F(\cdot)}$. For modal-incomplete cases, the hidden states corresponding to the absent modalities will be discarded.
The hidden states $H\subseteq \{h_1,h_2, ...,h_m\}$ are then fused.
\begin{equation}
    H_{f} = \mathcal{F}(H)
\end{equation}
The final loss function $\mathcal{L}(\cdot)$ and the corresponding gradient w.r.t. the shared parameters $\theta$ are computed between the fused representations and the target label.
\begin{equation}
    \mathcal{G}(H) = \nabla_\theta \mathcal{L}(\mathcal{F}(H), Y)
\end{equation}
The GMD method is applied to the gradients of two varied modal-incomplete representations $H_1$ and $H_2$ (e.g., the gradient of ${x_1, x_2}$ and ${x_2, x_3}$) to decouple the conflicts among them. 
As a result, the calibrated gradients of shared parameters can be combined and reduced without conflicts. The final gradient is then back-propagated through the shared backbone and individual modality encoders. 
\begin{equation}
    \hat{\mathcal{G}}(H_1), \hat{\mathcal{G}}(H_2) = \text{GMD}(\mathcal{G}(H_1), \mathcal{G}(H_2))
\end{equation}

\subsection{Modality Dominance in Missing Modality}
\label{modality_dependency}
Multimodal models tend to be dominated by some vital modalities while others remain under-optimized \cite{wu2022characterizing}.
On the one hand, modality dominance leads to under-optimization of unimodal ones, e.g. T1 of RFNet \cite{ding2021rfnet} performs worse than baseline in Table \ref{tab:dominance}. 
On the other hand, modality dominance suppresses the capability of modalities in multimodal settings.
As shown in Table \ref{tab:dominance}, methods like ACN and SMU-Net perform worse with multiple modalities than with individual modalities, indicating overly dependency on specific modalities.
Both cases have a significant impact on the robustness of missing modalities. 
Decoupling the multimodal model's dependency on dominating modalities is the key to addressing the missing-modality problem.

\subsection{Gradient-guided Modality Decoupling}
\label{GMD}
As discussed above, the modalities are coupled during training, leading to under-optimized modalities. {In the context of multimodal learning, determining the level of modality dominance can be challenging due to the joint optimization of modalities. When faced with missing modalities, a commonly used strategy is to improve the robustness of the multimodal model by training with modal-incomplete input, i.e., dropping some modalities during training. 
Similarly, we propose using the modal-incomplete input to explore the coupling of various modalities. 
}

\paragraph{Modal-incomplete gradient component}
When optimizing the multimodal framework with modal-incomplete subset $C_j$, i.e. some modalities are missing, the corresponding gradient $\mathcal{G}(C_j)$ can be viewed as the modal-incomplete component in the gradient space. Specially, when only one modality $x_m$ is presented, $\mathcal{G}(\{x_m\})$ is then considered the unimodal gradient component for modality $x_m$. 
This enables effective analysis of multimodal interactions and provides insights into the role of modalities in the learning process. For instance, by comparing $\mathcal{G}(\{x_1\})$ and $\mathcal{G}(\{x_2, x_3,...,x_m\})$, we can study the dominance and conflicts between modality ${x_1}$ and the rest of the modalities $\{x_2, x_3, ..., x_m\}$. 
As we investigate the modal-incomplete gradient components, we find that the modality dominance problem can be interpreted by ``gradient conflicts". For notational simplicity, we let $\mathcal{G}_j$ denote the gradient component for
modal-incomplete case $C_j$.

\begin{figure}[tb]
    \centering
    \includegraphics[width=0.45\textwidth]{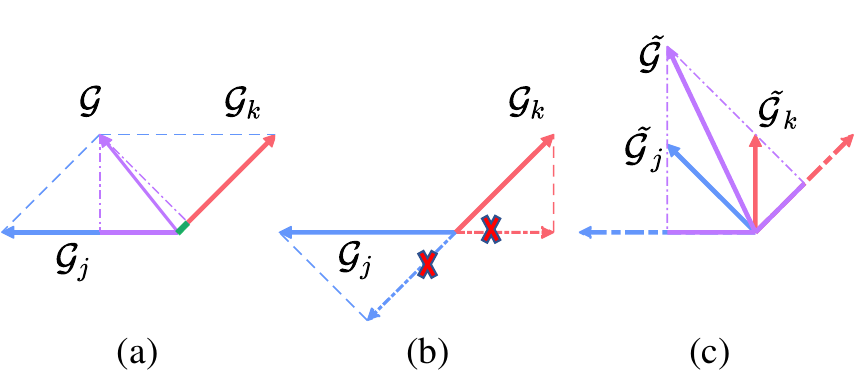}
    \caption{(a) demonstrates one gradient dominating another. The combined gradient $\mathcal{G}$ only has a small projected component at the direction of $\mathcal{G}_k$, highlighted in green; (b) shows the conflicting components to be removed, marked as \xmark; (c) visualizes the calibrated gradients. The corrected gradients are now of a similar scale and direction. Both modal-incomplete cases can be effectively updated without conflict during optimization.  }
    \label{fig:reduction}
\end{figure}

\paragraph{Gradient conflicts}
As shown in Figure \ref{fig:reduction} (a), by projecting the combined gradient to the direction of each modal-incomplete gradient component, we find that the effective update on different directions could be extremely imbalanced and biased towards the stronger case. 
To quantify this gradient conflict, we first compute the cosine similarity between two gradients,
$\mathcal{G}_j$ and $\mathcal{G}_k$, 
If $\mathcal{S}_{jk}$ is  positive, 
$\mathcal{G}_j$ and $\mathcal{G}_k$ share a common optimization direction, and both missing-modality cases can be well-optimized without gradient conflicts.
Otherwise, a negative  $\mathcal{S}_{jk}$ represents a 
significant direction difference (i.e., gradient conflicts) between $\mathcal{G}_j$ and $\mathcal{G}_k$. The optimization is at risk of being misled, leading to an inferior optimization. 

\begin{equation}
    \mathcal{S}_{jk} = \frac{\mathcal{G}_j\cdot \mathcal{G}_k}
                                              {||\mathcal{G}_j||||\mathcal{G}_k||}
\end{equation}

\begin{figure}[tb]
    \centering
   \includegraphics[width=.30\textwidth]{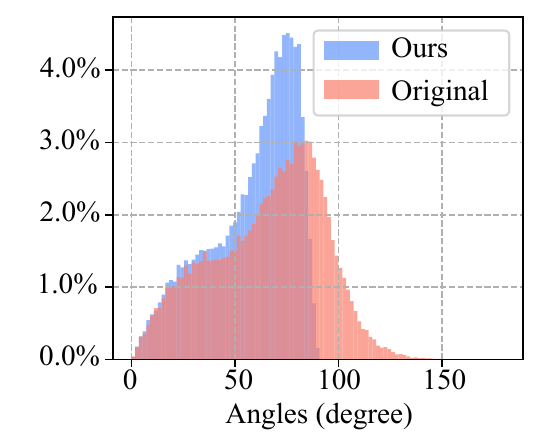}
    \caption{Gradients from modalities vary significantly in norm (std: $4\times10^4$). Previous works ignore gradient conflicts, leading to bias towards dominant gradients (red). Our GMD removes conflicting gradients and reduces angle differences, achieving more balanced optimization (blue).}
    \label{fig:modality_dominance}
\end{figure}

\paragraph{Gradient conflicts lead to modality dominance} 
Gradient conflicts indicate different modalities coupled with each other during training with modal-incomplete cases, leading to degraded performance. In particular, if there is one modal-incomplete input with a large gradient norm, it will dominate the optimization process. 
\begin{equation}
\begin{split}
    ||\mathcal{P}_{\mathcal{G}_k}^{\mathcal{G}_j}|| & = \frac{||\mathcal{G}_k\cdot \mathcal{G}_j||}{||\mathcal{G}_k||} \\
                                          & = ||\mathcal{S}_{jk}||||\mathcal{G}_j||
                                           \gg ||\mathcal{G}_k||
\end{split}
\end{equation}
We theoretically verify that a gradient with a significant norm dominates the final gradient $\mathcal{G}_{reduced}$.
With the assumption that $S_{jk} < 0$ and $||\mathcal{G}_j|| \gg ||\mathcal{G}_k||$ (a commonly exist phenomenon as shown in Figure \ref{fig:modality_dominance}), the projection of $\mathcal{G}_j$ onto the normal plane of $\mathcal{G}_k$ satisfies:
Therefore, the final reduced gradient between  $S_j$ and $S_k$ in a vanilla optimization process becomes:
\begin{equation}
\begin{split}
    \mathcal{G}_{reduced} & = \mathcal{G}_j + \mathcal{G}_k \\
                          & = (\mathcal{G}_j - \mathcal{P}_{\mathcal{G}_k}^{\mathcal{G}_j}) 
                              + (\mathcal{P}_{\mathcal{G}_k}^{\mathcal{G}_j} + \mathcal{G}_k) \\
                          & \approx (\mathcal{G}_j - \mathcal{P}_{\mathcal{G}_k}^{\mathcal{G}_j}) 
                              + \mathcal{P}_{\mathcal{G}_k}^{\mathcal{G}_j} \\
                          & = \mathcal{G}_j
\end{split}
\end{equation}
Thus, the reduced gradient for the final optimization process is at risk of being dominated by $\mathcal{G}_j$, leading to the coupling of modalities. 

\paragraph{Solution} To reduce modality dominance caused by conflicting gradients, we propose a Gradient-guided Modality Decoupling (GMD) method to
alter the gradients and eliminate the conflicting components when the gradient conflicts happen. 
Specifically,  when the cosine similarity $\mathcal{S}_{jk}$ is negative, we remove the projection of $\mathcal{G}_k$ onto $\mathcal{G}_j$, denoted as $\mathcal{P}_{\mathcal{G}_j}^{\mathcal{G}_k}$, from $\mathcal{G}_k$, leaving only the component orthogonal to $\mathcal{G}_j$. 
The same procedure is applied to $\mathcal{G}_j$.
\begin{equation}
    \mathcal{\Tilde{G}}_j = \mathcal{G}_j - \mathcal{P}_{\mathcal{G}_k}^{\mathcal{G}_j},\ \ \mathcal{\Tilde{G}}_k = \mathcal{G}_k - \mathcal{P}_{\mathcal{G}_j}^{\mathcal{G}_k}
\end{equation}
This essentially entails eliminating the competing gradient components among modalities, which reduces the harmful gradient interference between gradients of different modal-incomplete cases. As shown in Figure \ref{fig:reduction} (c), the corrected gradients can now be combined more efficiently without negative interplay. 
To extend to a multimodal framework with more than two missing-modality cases, we perform the same process on any pair of cases with conflicting components, i.e., negative cosine similarity. 
Our de-conflict process can be viewed as adaptively adjusting weights on conflicting gradients, preventing one from dominating the other,
\begin{equation}
\begin{split}
    \mathcal{\Tilde{G}} & 
                           = (1 - \frac{\mathcal{G}_k\cdot \mathcal{G}_j}{||\mathcal{G}_j||^2})\mathcal{G}_j + (1 - \frac{\mathcal{G}_k\cdot \mathcal{G}_j}{||\mathcal{G}_k||^2})\mathcal{G}_k
\end{split}
\label{eqn:gmd_weights}
\end{equation}
As shown in Figure \ref{fig:reduction} (c), the reduced gradient $\mathcal{\Tilde{G}}$ has projections on the original plane of $\mathcal{G}_j$ and $\mathcal{G}_k$ with a similar norm. Therefore, the missing-modality case $S_j$ and $S_k$ can both be effectively updated without bias. 
As shown in Figure \ref{fig:modality_dominance}, GMD adjusts the gradients and reduces the angle between most gradients to be within $90^\circ$,
which effectively resolves gradient conflicts and alleviates the modality dominance problem. 

  

\subsection{Dynamic Sharing (DS)}
\label{DS}
The existing multimodal models \cite{wang2021acn, azad2022smu} fuse different modalities' features or input data before producing the final prediction. Fusion operators, e.g., concatenation, require modal-incomplete input to be completed by masking values \cite{dorent2019hetero} or reconstructed values \cite{latent,zhou2021conditional}, which may introduce noises to fusion operators, causing unexpected behavior of models.

{The observation presented leads us to propose the Dynamic Sharing (DS) framework, as depicted in Figure \ref{fig:ds_framework}. This framework has three main components: modality-specific encoders, a shared backbone, and a feature fusion mechanism.
Each modality, represented by $X_m$, is initially processed by its modality-specific encoder $f_{\theta_m}$ to a hidden representation, $H_m \in \mathbb{R}^h$, with a dimension of $h$. $H_m$ is then separately fed into the shared backbone $f_{\theta_s}$, to obtain modality-specific features, $O_m$.
Finally, different modality-specific features are combined by a flexible feature fusion operator independent of the number of inputs, such as average pooling, to produce the final representation.} 
With the aid of modality-specific components and flexible feature fusion, we can adaptively process the modal-incomplete data without being affected by corrupted modality.  
Complementary information among modalities could be utilized through shared parameters implicitly \cite{chang2019domain}.

\subsubsection{Correlation between GMD and DS}
The GMD combined with the DS framework collectively offers a comprehensive solution throughout the network optimization process.
Specifically, DS and GMD target the missing modality issue in the forward and backward passes, respectively. DS allows clean propagation of only the available modalities, avoiding misleading fusion when modalities are missing. This prevents ``corrupted" gradients from ever being calculated in the first place.
Meanwhile, GMD acts as a safety net during backpropagation, which provides a robust decoupling mechanism to resolve the conflicts in the gradients. 
The result is a harmonious interplay between DS and GMD. 
This joint effort at both the input and output of the network training process allows DS and GMD to effectively and efficiently enhance the missing modality robustness.
\begin{table*}[tbh]
\centering
\fontsize{9pt}{9pt}\selectfont
\begin{tabular}{c|l|w{c}{1.2em}w{c}{1.2em}w{c}{1.2em}w{c}{1.2em}w{c}{1.9em}w{c}{1.9em}w{c}{1.9em}w{c}{1.9em}w{c}{1.9em}w{c}{1.9em}w{c}{1.2em}w{c}{1.2em}w{c}{1.2em}w{c}{1.2em}w{c}{1.2em}|c}
\toprule
Task                          &  Methods  & Fl.         & T2            & T1c          & T1  &      \underline{T2,Fl.}      & \underline{T1c,Fl.}    & \underline{T1c,T2}       & \underline{T1,Fl.}     & \underline{T1,T2}         & \underline{T1,T1c}     & $\sim$T1      & $\sim$T1c    & $\sim$T2      & $\sim$Fl.   & Full          & Avg.   \\ \midrule
\multirow{9}{*}{WT}& Baseline      & 81.4           & 70.1             & 78.3          & 79.6          & 75.4          & 83.7           & 74.0           & 84.4          & 73.0           & \textbf{81.9} & 84.7          & 84.6          & 85.4           & 83.3          & 86.8           & 81.6          \\
 & U-HeMIS       & 79.9           & 79.2             & 58.5          & 54.3          & 86.0          & 83.3           & 81.0           & 83.9          & 80.8           & 63.8          & 87.0          & 87.0          & 85.1           & 82.1          & 87.6           & 77.9          \\
 & HVED          & 82.1           & 80.9             & 62.4          & 52.4          & 87.5          & 85.5           & 82.7           & 84.3          & 82.2           & 66.8          & 88.6          & 88.0          & 86.2           & 83.3          & 88.8           & 79.2          \\
 & ACN           & 87.3           & 85.6             & {80.5}        & {79.3}        & 87.8          & 88.3           & 86.4           & 87.5          & 85.5           & 80.1          & 88.3          & 88.4          & 89.0           & 86.9          & 89.2           & 86.1          \\
 & RFNet         & 87.3           & {86.1}           & 76.8          & 77.2          & 81.8          & 89.9           & 87.7           & 81.1          & 87.7           & 81.1          & 90.7          & 90.6          & 90.7           & {88.3}        & {91.1}         & 86.5          \\
 & SMU-Net       & 87.5           & 85.7             & 80.3          & 78.6          & 87.9          & 88.4           & 86.1           & 87.3          & 85.6           & 80.3          & 88.2          & 88.3          & 88.2           & 86.5          & 88.9           & 85.8          \\
 & OGM-GE        & 72.4           & 61.4             & 71.2          & 30.7          & 77.5          & 78.7           & 73.7           & 67.2          & 57.6           & 58.5          & 81.3          & 71.4          & 71.6           & 64.4          & 74.7           & 66.6          \\
 & \textbf{Ours} & \textbf{89.3}  & \textbf{87.0}    & \textbf{80.8} & \textbf{80.3} & \textbf{90.7} & \textbf{90.6}  & \textbf{88.5}  & \textbf{90.2} & \textbf{88.2}  & \textbf{83.3} & \textbf{91.3} & \textbf{91.0} & \textbf{90.8}  & \textbf{89.0} & \textbf{91.3}  & \textbf{87.9} \\ \midrule
\multirow{9}{*}{TC}  & Baseline      & 58.3           & 56.3             & 80.0          & 71.0          & 61.8          & 74.1           & 71.1           & 68.9          & 65.6           & 79.5          & 71.6          & 67.9          & 76.2           & 74.9          & 75.4           & 70.2          \\
 & U-HeMIS       & 49.8           & 50.5             & 58.5          & 37.9          & 58.7          & 67.6           & 69.1           & 56.7          & 53.4           & 64.0          & 72.2          & 61.5          & 70.7           & 70.7          & 73.4           & 60.5          \\
 & HVED          & 50.4           & 54.1             & 66.7          & 37.2          & 59.7          & 72.9           & 73.7           & 55.3          & 57.2           & 69.7          & 75.6          & 61.5          & 74.2           & 75.3          & 76.4           & 63.5          \\
 & ACN           & 67.7           & 67.9             & \textbf{84.2} & {71.2}        & 71.6          & 83.4           & 84.4           & 71.3          & 73.3           & \textbf{84.6} & 82.9          & 67.9          & 84.3           & 84.7          & 85.2           & 77.3          \\
 & RFNet         & 69.2           & 71.0             & 81.5          & 66.0          & 74.1         & \textbf{84.7} & 83.5          & 73.1         & 73.1          & 83.4          & \textbf{85.0} & 75.2          & \textbf{85.1}  & 83.5          & 85.2           & 78.0          \\
 & SMU-Net       & 71.8           & 67.2             & 84.1          & 69.5          & 71.2          & 84.1           & \textbf{85.0 } & 71.2          & 73.5           & 84.4          & 82.5          & 67.9          & 84.2           & 84.4          & \textbf{87.3}  & 77.7          \\
 & OGM-GE        & 53.7           & 45.3             & 62.3          & 29.9          & 58.0          & 64.7           & 62.4           & 50.9          & 42.3           & 57.4          & 64.9          & 49.4          & 60.4           & 55.9          & 59.2           & 53.5          \\
 & \textbf{Ours} & \textbf{74.1 } & \textbf{75.4}    & 82.4          & \textbf{73.2} & \textbf{77.8} & 83.9           & 83.9           & \textbf{77.1} & \textbf{78.3}  & 83.4          & 84.7          & \textbf{79.0} & 84.8           & \textbf{84.7} & 84.9           & \textbf{80.4} \\
                                 \midrule
\multirow{9}{*}{ET}  & Baseline      & 35.4           & 42.2             & 81.1          & 37.6          & 44.6          & 63.2           & 66.4           & 39.7          & 44.9           & 70.9          & 65.0          & 49.3          & 58.2           & 64.6          & 61.6           & 55.0          \\
 & U-HeMIS       & 24.9           & 23.3             & 60.8          & 12.4          & 28.0          & 68.0           & 68.6           & 29.0          & 28.3           & 65.3          & 69.7          & 33.4          & 69.9           & 69.7          & 70.8           & 48.3          \\
 & HVED          & 24.8           & 30.8             & 65.5          & 13.7          & 34.6          & 70.3           & 70.2           & 24.2          & 30.7           & 67.0          & 71.2          & 34.1          & 71.1           & 71.1          & 71.7           & 50.4          \\
 & ACN           & 42.8           & 43.0             & 78.1          & 41.5          & 46.0          & 77.5           & 75.7           & 43.7          & 47.4           & 75.2          & 76.0          & 42.1          & 76.2           & 76.1          & 77.1           & 61.4          \\
 & RFNet         & 38.2           & 46.3             & 74.9          & 37.3          & 49.3         & 76.7          & 75.9          & 41.0         & 45.7          & 78.0         & 77.1          & 49.9          & 76.8           & 76.8          & 78.0           & 61.7          \\
 & SMU-Net       & 46.1           & 43.1             & 78.3          & \textbf{42.8} & 46.0          & 77.3           & 75.7           & 44.0          & 47.7           & 77.3          & 75.4          & 43.1          & 76.2           & 76.2          & 79.3           & 62.3          \\
 & OGM-GE        & 29.5           & 24.5             & 73.9          & 12.0          & 28.5          & 51.9           & 53.4           & 28.3          & 20.7           & 46.8          & 45.0          & 26.1          & 48.8           & 40.9          & 43.1           & 38.2          \\
 & \textbf{Ours} & \textbf{47.4}  & \textbf{57.0   } & \textbf{84.5} & 42.2          & \textbf{59.0} & \textbf{85.1}  & \textbf{84.7}  & \textbf{51.1} & \textbf{57.9 } & \textbf{84.4} & \textbf{84.5} & \textbf{59.2} & \textbf{84.6 } & \textbf{84.2} & \textbf{84.0 } & \textbf{69.7}    \\
                                 \bottomrule
\end{tabular}
\caption{Experimental results on BraTS 2018 dataset measured by DSC($\uparrow$). 
 Flair, T2, T1CE and T1 are 4 MRI modalities. $\sim (\cdot)\ $indicates the modality is absent. Our proposed method achieves the best average performance on all segmentation tasks under various modal-incomplete cases and full-modal scenarios.
Our method significantly improves model robustness, especially on the most challenging task ET.}
\label{tab:brats}
\end{table*}
\subsection{Complexity Analysis}
\label{method_complexity}
As described above, the proposed GMD approach utilizes missing modality cases during training to resolve modality dominance. When all modal-incomplete cases are considered in each iteration, it requires fusing $O(2^M)$ combinations of modalities for a dataset with M modalities. 
To improve the efficiency, instead of fusing all possible modality combinations, we sample a subset of $k$ combinations at each training iteration. $k$ is set to balance efficiency and coverage (e.g. k=5). This allows approximating GMD with much fewer fused representations and backward gradient computation. Our experiments show that reducing the GMD combinations from $O(2^m)$ to $O(km)$ greatly improves scalability. 
In addition, the approximated GMD achieves similar performance to the full version, demonstrating its effectiveness.

\section{Experiments}
\subsection{Datasets}
\paragraph{Brain tumor segmentation (BraTS)}
Multimodal Brain Tumor Segmentation Challenge (BraTS 2018) \cite{brats} dataset is comprised of 285 multi-contrast MRI scans with four MRI modalities: (1) native (T1), (2) post-contrast T1-weighted (T1CE),(3) T2-weighted (T2), and (4) T2 Fluid Attenuated Inversion Recovery (Flair). 
\paragraph{Multimodal sentiment analysis (MSA)}
The CMU-MOSI dataset \cite{zadeh2016mosi} is a popular benchmark for Multimodal Sentiment Analysis (MSA). 
The dataset is a collection of YouTube monologues where speakers express their opinions on various topics.
The CMU-MOSEI dataset \cite{zadeh2018mosei} is an enhanced version of CMU-MOSI with a more extensive collection of utterances and improved variety in samples, speakers, and topics. 

\begin{table}[htb]
\centering
\fontsize{9pt}{9pt}\selectfont
\begin{tabular}{cc|cccc|c}
\toprule
\multicolumn{2}{c}{Methods} & \multicolumn{4}{c}{\# of Missing Modality} &         \\
\midrule
DS      &  GMD     & 3          & 2         & 1         & 0          & Avg. \\
\midrule
        &          & 49.1       & 54.9     &  59.3     &  61.6     &  56.2  \\
\cmark  &          & 46.9      & 61.8     & 66.6     & 70.8     & 61.5    \\
\cmark  &  \cmark  & \textbf{54.8}      & \textbf{69.1}     & \textbf{76.8}     & \textbf{82.1}      & \textbf{70.7}   \\
\bottomrule
\end{tabular}
\caption{Ablation studies on our proposed approaches. The performance is measured by DICE ($\uparrow$) on the BraTS dataset. Both DS and GMD significantly contribute to the robustness of our framework.}
\label{tab:ablation}
\end{table}

\subsection{Implementation Details}

We evaluate the performance of all presented models following conventional setting \cite{tsai2018learning}, where training modalities are all available. After training, models are tested against all missing-modality scenarios, i.e. missing $1, 2, ..., m-1$ modalities. For evaluations on the BraTS dataset, the Dice similarity coefficient (DSC $\uparrow$) is used. For experiments conducted on MSA, we follow the settings adopted in GMC \cite{poklukar2022geometric} and report the mean accuracy (Acc) and F1 score (F1) missing-modality and full-modality scenarios. Experiments are conducted on 4$\times$RTX3090 with 24 GB memory. The reported results are averaged over 5 runs.

\paragraph{Missing modality baseline} 
Training each modality with a dedicated network is an intuitive way to demonstrate the capability of individual modalities on various tasks. In addition, since no interaction between modalities is involved, an ensemble of unimodal networks provides a multimodal fusion baseline for multimodal performance.
\subsection{Performance on BraTS and MSA}
\label{brats_analysis}

\begin{table}[tbh]
\centering
\fontsize{9pt}{9pt}\selectfont
\begin{tabular}{cl|w{c}{1em}w{c}{1em}w{c}{1em}w{c}{1em}w{c}{1em}w{c}{1em}w{c}{1em}|w{c}{1.8em}}
\toprule
                                            &          & \multicolumn{7}{c|}{Available Modalities}                                                                     & \multicolumn{1}{l}{} \\ \midrule
\multicolumn{1}{c|}{}                       & Meth.   & l             & a             & v             & a,v           & l,a           & l,v           & Full          & Avg. $\uparrow$    \\ \midrule
\multicolumn{1}{c|}{\multirow{6}{*}{\rotatebox{270}{MOSI}}}  & CPM.  & 77.3          & 54.3          & 55.8          & 53.7          & 75.6          & 74.8          & 70.6          & 66.0                 \\
\multicolumn{1}{c|}{}                       & GMC      & 75.8          & \textbf{57.6} & 55.6          & 58.1          & 73.5          & 73.2          & 77.6          & 67.3                 \\
\multicolumn{1}{c|}{}                       & MMIN     & 84.0          & 54.6          & 55.2          & 59.5          & 83.7          & 84.3          & 84.1          & 72.2                 \\
\multicolumn{1}{c|}{}                       & GCN.    & 82.5          & 51.7          & \textbf{57.3} & \textbf{60.2} & 84.1          & 82.2          & 82.2          & 71.5                 \\
\multicolumn{1}{c|}{}                       & Basel. & \textbf{84.9} & 51.4          & 54.4          & 58.4          & 81.7          & 81.9          & 80.8          & 70.5                 \\
\multicolumn{1}{c|}{}                       & Ours     & \textbf{84.9} & 56.9          & 56.7          & \textbf{60.2} & \textbf{85.4} & \textbf{85.2} & \textbf{86.0} & \textbf{73.6}        \\  \midrule
\multicolumn{1}{c|}{\multirow{6}{*}{\rotatebox{270}{MOSEI}}} & CPM.  & 77.7          & 65.8          & 59.1          & 55.0          & 79.3          & 78.9          & 79.1          & 70.7                 \\
\multicolumn{1}{c|}{}                       & GMC      & 76.7          & 65.0          & 63.8          & 66.8          & 75.0          & 75.6          & 80.0          & 71.8                 \\
\multicolumn{1}{c|}{}                       & MMIN     & 85.8          & 69.2          & 65.4          & 70.1          & 85.6          & 85.6          & 85.5          & 78.2                 \\
\multicolumn{1}{c|}{}                       & GCN.    & 86.4          & 67.1          & 66.5          & 66.5          & 85.9          & 86.4          & 86.4          & 77.9                 \\
\multicolumn{1}{c|}{}                       & Basel. & 86.1          & 71.4          & 68.8          & 72.2          & 84.5          & 84.5          & 82.3          & 78.6                 \\
\multicolumn{1}{c|}{}                       & Ours     & \textbf{86.4} & \textbf{72.3} & \textbf{70.4} & \textbf{72.6} & \textbf{86.7} & \textbf{87.1} & \textbf{87.1} & \textbf{80.4}        \\ \bottomrule
\end{tabular}
\caption{Experimental results on CMU-MOSEI and CMU-MOSI dataset, measured by prediction accuracy (Acc) ($\uparrow$). l, a,v represents language, audio and video modalities, respectively. Among the compared methods, our proposed approach achieves the best overall performance.}
\label{tab:mosi}
\end{table}

\begin{table}[tbh]
\centering
\fontsize{9pt}{9pt}\selectfont
\begin{tabular}{l|cccc}
\toprule
        & \multicolumn{4}{c}{Modal-incomplete cases used in training}                                        \\
        \midrule
Task    & missing 1 & missing 2 & missing 3 & all cases \\
\midrule
WT      & 82.9      & 86.8      & 85.8      & 88.2      \\
TC      & 75.9      & 80.1      & 80.5      & 80.5      \\
ET      & 65.3      & 68.3      & 60.4      & 70.0      \\
\midrule
Average & 74.7               & 78.4               & 75.6               & 79.5               \\
\bottomrule
\end{tabular}
\caption{Missing modality performance of GMD using a subset of modal-incomplete cases on BraTS dataset measured by DSC ($\uparrow$). With only cases of missing 1, 2, or 3 modalities  used during training, GMD can still achieve comparable performance to the full version.}
\label{tab:cases}
\end{table}

\label{brats18_results}
%
\paragraph{Performance on BraTS}
Compared to existing methods, our method is least impacted by missing-modal data overall. In addition, our method achieves superior performance on the most challenging ET task, with a solid improvement of $5.3\%$ over the previous method SMU-Net.
With modal-complete data, our method achieves comparable performance to the existing method,
which proves that our approach can improve the missing-modality robustness without sacrificing the full-modal performance. 

\paragraph{Analysis of modality dominance}
The results obtained on BraTS 2018, shown in Table \ref{tab:brats}, validate our analysis and findings regarding the modality dominance problem. In Table \ref{tab:brats}, most existing approaches exhibit degraded performance multimodal performance compared to their unimodal performance. Moreover, in some cases, the unimodal performance is not fully optimized compared to the baseline. In contrast, our proposed method achieves the best unimodal performance for 3 out of 4 modalities and the highest multimodal performance. This demonstrates that our gradient-guided modality decoupling approach effectively addresses the modality dominance problem by properly balancing and optimizing all modalities simultaneously. 


\paragraph{Performance on MSA}

GMC employs a strong baseline Multimodal Transformer \cite{tsai2019multimodal} for better multimodal performance. However, our method achieves comparable multimodal performance to GMC without additional multimodal branch. Moreover, our approach improves on top of the baseline method by utilizing complementary information among different modalities, achieving the best average performance of all scenarios. 

\subsection{Ablation Study} 
\label{ablation}

\begin{figure}[t]
    \centering
    \includegraphics[width=.43\textwidth]{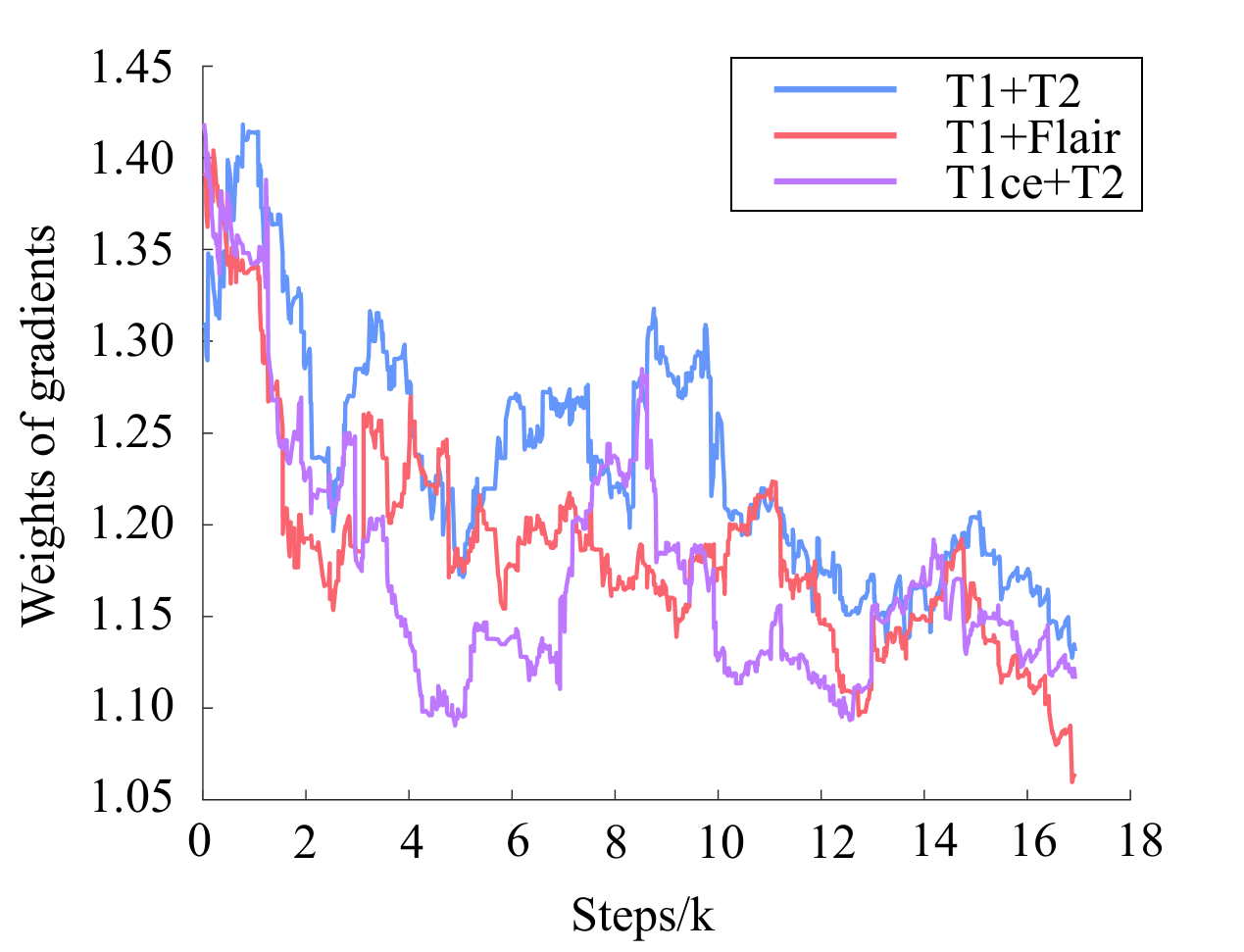}
    \caption{Visualization of weights given by GMD on different modal-incomplete cases, smoothed with a window size of 100 steps. For cases with T1CE, the dominant modality, presented, its gradient (purple line) is relatively suppressed to allow for optimization of cases with only weak modalities (red and blue line).}
    \label{fig:coefs}
\end{figure}

\paragraph{Analysis on GMD}

As demonstrated in equation (\ref{eqn:gmd_weights}), the de-conflict process of GMD can be interpreted as adaptively assigning weights to gradients.
As shown in Figure \ref{fig:coefs}, the weights given by GMD vary in different modal-incomplete cases.
For the dominating modality (T1CE), its gradient is dynamically adjusted and suppressed (relative to other cases) to allow cases with only weak modalities to be well-optimized. 
The adaptive adjustment of gradient weights proves that GMD can decouple the modality dominance by identifying and resolving conflicts between gradients.


\paragraph{Analysis on DS framework}
Dynamic Sharing (DS) framework, compared to the vanilla multimodal framework, the flexibility is greatly improved. 
For modal-incomplete cases, DS does not require the imputation of missing modalities. 
Instead, only the layers related to presented modalities are used. 
Thus, the framework is not affected by masking values contrary to the existing method. 
As a result of shared parameters, the performance of DS on multimodal input is significantly improved compared to the missing-modality baseline, as shown in Table \ref{tab:ablation}.
\paragraph{Efficiency of GMD}
Table \ref{tab:cases} demonstrates how the choice of modal-incomplete cases used for GMD affects the performance.
Missing 1 modality cases can not fully guarantee strong unimodal performance, as a single modality may still dominate the others. On the other hand, missing 3 modalities sacrifices the multimodal fusion capability. In contrast, the missing 2 cases induce balanced gradient conflicts between modalities, exposing the core issues of dominance while retaining sufficient uni- and multi-modal representations. 

%
\section{Conclusion}
In this paper, we identify the modality dominance problem in the context of missing modality that leads to degraded performance on modal-incomplete data. Based on theoretical analysis and experiments, we reformulate the modality dominance problem into gradient conflicts during optimization. We present GMD--a gradient-guided modality decoupling method to address modality dependency and achieve robustness to the missing-modality problem. In addition, to flexibly deal with modal-complete, we propose a parameter-efficient Dynamic Sharing (DS) architecture. Experiments across multiple multimodal datasets demonstrate the superior robustness of our method. 

\section*{Acknowledgments}
This work is supported in part by National Key Research and Development Program of China (2021YFF1200800) and National Natural Science Foundation of China (Grant No. 62276121); and Guangdong Basic and Applied Basic Research Foundation (Grant No. 2022A1515110573).
\bibliography{aaai24}

\end{document}